\documentclass[letterpaper, 10 pt, conference]{ieeeconf}  

\IEEEoverridecommandlockouts                              

\usepackage{amsmath,amssymb,amsfonts}
\usepackage{algorithmic}
\usepackage{graphicx}
\usepackage{textcomp}
\usepackage{xcolor}
\usepackage[nolist]{acronym}
\usepackage{siunitx}
\usepackage{hhline}
\usepackage{multirow}
\usepackage{bm}
\usepackage{pifont}
\usepackage[hidelinks]{hyperref}
\usepackage{cleveref}
\usepackage{color,soul}
\usepackage{array}
\usepackage{booktabs}
\usepackage{tabularray}
\UseTblrLibrary{booktabs}
\usepackage[nolist]{acronym}
\usepackage{flushend}   
\usepackage{wrapfig, lipsum}
\usepackage{bbm}
\usepackage{subcaption}
\usepackage{dblfloatfix}

\usepackage[dvipsnames]{xcolor}
\usepackage{tikz}
\usepackage{tkz-tab}
\usetikzlibrary{decorations.pathreplacing}
\usetikzlibrary{fadings}
\usetikzlibrary{calc,positioning,fit,backgrounds}
\usetikzlibrary{arrows,positioning,shapes.geometric}
\usepackage{tikz-3dplot}
\usepackage{tkz-euclide}
\usepackage{tkz-graph}

\usepackage{pgfplots}
\usepackage{pgfplotstable}
\pgfplotsset{compat=1.9}
\usepgfplotslibrary{external}

\begin{acronym}
    
    \acro{av}[AV]{Autonomous Vehicle}
	\acroplural{av}[AVs]{Autonomous Vehicles} 
    \acro{bev}[BEV]{bird's-eye-view}
	\acro{sotif}[SOTIF]{Safety Of The Intended Functionality}
	\acro{fov}[FOV]{Field of View}
    \acro{oem}[OEM]{Original Equipment Manufacturer}
    \acro{saf}[SAF]{Safety Assurance Framework}
    \acro{ccam}[CCAM]{Cooperative Connected and Automated Mobility}
    \acro{odd}[ODD]{Operational Design Domain}    
    \acro{wmm}[WMM]{Without Motion Model}
    \acro{hms}[HMS]{Hybrid Multi Shot}
    \acro{hss}[HSS]{Hybrid Single Shot}
    \acro{cv}[CV]{Constant Velocity}
    \acro{ctra}[CTRA]{Constant Turn Rate and Acceleration}
    \acro{pem}[PEM]{Perception Error Model}
    
	\acro{ai}[AI]{Artificial Intelligence}
	\acro{ml}[ML]{Machine Learning}
	\acro{dl}[DL]{Deep Learning} 
	\acro{dnn}[DNN]{Deep Neural Network} 
	\acroplural{dnn}[DNNs]{Deep Neural Networks}
    \acro{mlp}[MLP]{Multi-Layer Perceptron}
	\acro{cnn}[CNN]{Convolutional Neural Network} 
	\acroplural{cnn}[CNNs]{Convolutional Neural Networks}
	\acro{lstm}[LSTM]{Long Short-Term Memory} 
	\acro{rnn}[RNN]{Recurrent Neural Network}
    \acro{dkm}[DKM]{Deep Kinematic Model}
    \acro{gnn}[GNN]{Graph Neural Network}
	\acroplural{gnn}[GNNs]{Graph Neural Networks} 
    \acro{gft}[GFT]{Graph Fourier Transform}
    \acro{ema}[EMA]{Exponential Moving Average}
    \acro{amp}[AMP]{Automatic Mixed Precision}
    \acro{jepa}[JEPA]{Joint Embedding Predictive Architecture}
    \acro{ssl}[SSL]{Self-Supervised Learning}
    \acro{flop}[FLOP]{Floating-Point OPeration}
    \acro{flops}[FLOPS]{Floating-Point OPerations per Second}
    \acroplural{flop}[FLOPs]{Floating-Point OPerations}
    \acro{vos}[VOS]{Virtual Outlier Synthesis}
    \acro{lsvos}[LS-VOS]{Latent Space Virtual Outlier Synthesis}
    
    \acro{ismr}[ISMR]{In-Service Monitoring and Reporting}
    \acro{smf}[SMF]{Safety Monitoring Framework}
    \acro{ood}[OOD]{Out-Of-Distribution}  
    \acro{kf}[KF]{Kalman-Filter}
    \acro{mot}[MOT]{Multi-Object Tracking}
    \acro{gmm}[GMM]{Gaussian Mixture Model}
	\acroplural{gmm}[GMMs]{Gaussian Mixture Models}
    \acro{dbscan}[DBSCAN]{Density-Based Spatial Clustering of Applications with Noise}
    \acro{ocsvm}[OCSVM]{One Class Support Vector Machine}
    \acro{abod}[ABOD]{Angle-Based Outlier Detection}
    \acro{copod}[COPOD]{Copula-Based Outlier Detection}
    \acro{lof}[LOF]{Local Outlier Factor}
    \acro{pca}[PCA]{Principal Component Analysis}
    \acro{tsne}[t-SNE]{t-Distributed Stochastic Neighbor Embedding}
    \acro{umap}[UMAP]{Uniform Manifold Approximation and Projection}
    \acro{mse}[MSE]{Mean Squared Error}
    \acro{giou}[gIoU]{generalized Intersection over Union}
    \acro{auc}[AUC]{Area Under the Curve}
    \acro{roc}[ROC]{Receiver Operating Characteristics}
    \acro{auroc}[AUROC]{Area Under the Receiver Operating Characteristic curve}
    \acro{fpr}[FPR]{False Positive Rate}
    \acro{tpr}[TPR]{True Positive Rate}
    \acro{mcc}[MCC]{Matthews Correlation Coefficient}
    \acro{fde}[FDE]{Final Displacement Error}
    \acro{ade}[ADE]{Average Displacement Error}

\end{acronym}

 \hypersetup{
     pdftitle={Online Monitoring Framework for Automotive Time Series Data using JEPA Embeddings},
     pdfdisplaydoctitle={true},
     pdfauthor={Alexander Fertig, Karthikeyan Chandra Sekaran, Lakshman Balasubramanian and Michael Botsch},
     pdfsubject={Embedding Method for Automotive Anomaly Detection},
     pdfkeywords={Autonomous Driving, Safety, Representation Learning, Anomaly Detection, SOTIF, ISO 21446, ISMR, IV, Deep Learning, Machine Learning, DL, ML, Latent Space, Self-supervised Learning, SSL, JEPA, Joint Embedding Predictive Architecture},
 }

\newcommand{\eg}{e.\,g., }      
\newcommand{\ie}{i.\,e., }      

\newcommand{\getGitHubURL}{\url{https://github.com/MB-Team-THI/jepa_online_monitoring_framework}}

\makeatletter
\IEEEtriggercmd{\reset@font\normalfont\footnotesize}
\makeatother
\IEEEtriggeratref{1}

\newcommand{\getNumObjectLiDARTrain}{18,253} 
\newcommand{\getNumObjectLiDARTest}{3,841} 

\newcommand{\getEpochsPretraining}{250 }

\newcommand{\getLearningRate}{$3 \times 10^{-5}$}

\newcommand{\getNumMasksPretraining}{4}

\newcommand{\getEncoderNumberHead}{10 }
\newcommand{\getEncoderDepth}{5}
\newcommand{\getEncoderLatentDim}{32}
\newcommand{\getEncoderNumLayerMLP}{3}
\newcommand{\getEncoderNumParams}{$4.5 \times 10^5$ }
\newcommand{\getEncoderEMA}{0.99 }

\newcommand{\getPredictorNumberHead}{4 }
\newcommand{\getPredictorDepth}{3}

\newcommand{\getPredictorNumParams}{$5.1 \times 10^4$}

\newcommand{\getNumFLOPs}{\mbox{$13.7 \times 10^{6}$} }
\newcommand{\getGMMComps}{5 }
\newcommand{\getLOFnumNeighbor}{15 }

\newcommand{\getNumbRunsAnomalyDetectionPerformance}{5 }

\newcommand\copyrighttext{%
  \footnotesize \textcopyright 
  2026 IEEE. Personal use of this material is permitted. Permission from IEEE must be obtained for all other uses, in any current or future media, including reprinting/republishing this material for advertising or promotional purposes, creating new collective works, for resale or redistribution to servers or lists, or reuse of any copyrighted component of this work in other works.
  DOI: \href{https://doi.org/10.1109/IV66570.2026.11624095}{10.1109/IV66570.2026.11624095}
  }
  
\newcommand\copyrightnotice{%
\begin{tikzpicture}[remember picture,overlay]
\node[anchor=south,yshift=10pt] at (current page.south) {\fbox{\parbox{\dimexpr\textwidth-\fboxsep-\fboxrule\relax}{\copyrighttext}}};
\end{tikzpicture}%
}

\title{\LARGE \bf
Online Monitoring Framework for Automotive\\
Time Series Data using JEPA Embeddings
}

\author{Alexander Fertig$^{1}$, Karthikeyan Chandra Sekaran$^{2}$, Lakshman Balasubramanian$^{3}$ and Michael Botsch$^{1,2}$
    \thanks{$^{1}$Technische Hochschule Ingolstadt, AImotion Bavaria, Esplanade 10, 85049 Ingolstadt, Germany, {\tt\small firstname.lastname@thi.de}}%
    \thanks{$^{2}$Technische Hochschule Ingolstadt, Research Center CARISSMA, Esplanade 10, 85049 Ingolstadt, Germany}%
    \thanks{$^{3}${\tt\small balasubramanianlakshman@gmail.com}}
}

\begin{document}

\bstctlcite{IEEEexample:BSTcontrol}

\maketitle
\copyrightnotice
\thispagestyle{empty}
\pagestyle{empty}

\begin{abstract}

As autonomous vehicles are rolled out, measures must be taken to ensure their safe operation.
In order to supervise a system that is already in operation, monitoring frameworks are frequently employed.
These run continuously online in the background, supervising the system status and recording anomalies.
This work proposes an online monitoring framework to detect anomalies in object state representations.
Thereby, a key challenge is creating a framework for anomaly detection without anomaly labels, which are usually unavailable for unknown anomalies.
To address this issue, this work applies a self-supervised embedding method to translate object data into a latent representation space.
For this, a JEPA-based self-supervised prediction task is constructed, allowing training without anomaly labels and the creation of rich object embeddings.
The resulting expressive JEPA embeddings serve as input for established anomaly detection methods, in order to identify anomalies within object state representations.
This framework is particularly useful for applications in real-world environments, where new or unknown anomalies may occur during operation for which there are no labels available.
Experiments performed on the publicly available, real-world nuScenes dataset illustrate the framework's capabilities.

\end{abstract}

\section{Introduction} 
\label{ch:introduction}


The number of vehicles with highly automated driving functions or \acp{av} on the roads is increasing, with this figure set to rise further in the coming years.
Thereby, safety-critical driving tasks are being transferred to autonomous systems, requiring a high level of safety assurance.
However, ensuring such safety requirements is extremely challenging, as \acp{av} can encounter a multitude of different scenarios and situations in real-world environments that are difficult to represent in usually small test datasets \cite{Bogdoll2025}.
The standard \mbox{ISO 21448 \cite{ISO21448},} \ac{sotif}, offers a remedy in the form of guidelines for the validation and verification processes of \acp{av}.
In \ac{sotif} the period after approval, when customers drive the \ac{av} on public roads is referred to as \textit{operation phase}.
\ac{sotif} prescribes monitoring activities during this operation phase in order to identify unknown and potentially hazardous situations \cite{ISO21448}, which are not or insufficiently represented in the validation and verification process.
In order to detect these unknown or overlooked risks, monitoring frameworks are applied, which observe the system and aim to identify anomalies in the system behavior.


\begin{figure}
   \centering
   \includegraphics[width=0.485\textwidth]{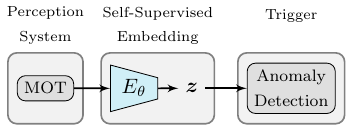}
   \caption{
        Framework overview:
        In the perception system \acf{mot} is performed to obtain object state estimations, which are encoded by $E_{\theta}$ into latent embeddings $\bm{z}$ and used for anomaly detection.
    }
    \label{fig:fig_framwork_overview}
    \vspace{-0.6cm}
\end{figure}
This work introduces an online monitoring framework to detect anomalies in automotive time series data.
As depicted in \Cref{fig:fig_framwork_overview}, the framework consists of three main components.
First, the perception system processes the vehicle's environment by performing \acf{mot} on  sensor data from arbitrary sensor(s), to create object state representations in the form of multivariate time series. 
Second, a meaningful latent representation space is shaped through a self-supervised training procedure based on a \acf{jepa} \cite{LeCun2022}. 
Thereby, each object state representation is embedded into a latent representation space.
Third, anomaly detection is performed on these resulting latent embeddings to identify anomalous samples within the object state representation.
The main contributions are:
\begin{itemize}
    \item Development of an online monitoring framework using a JEPA-based self-supervised embedding method to identify anomalies within multivariate time series data.
    \item A label-free training procedure realized by the self-supervised objective of the embedding method.
    \item Design of the monitoring framework for online application in \ac{sotif}'s operation phase (ISO 21448), for improving safety and online validation process of \acp{av}.
    \item Experiments on the publicly available real-world nuScenes dataset show the potential of the proposed framework. 
    The GitHub repository is publicly available:
    \small
    \getGitHubURL
\end{itemize}
    
\section{Related Work} 
\label{ch:related_work}
The \acf{jepa}, first introduced in \cite{LeCun2022}, represents a non-generative \ac{ssl} paradigm that enables the learning of rich representations without the need for explicit labels. 
The core idea is to train an encoder to process an \mbox{input $x$ }and predict the latent \mbox{representation $\hat{s}_y$} of the \mbox{target $y$}.
This prediction is supported by a predictor network that is usually conditioned on an additional latent variable $z$.
A key advantage of \ac{jepa} is that it focuses on a prediction within an abstract representation space.
This allows the model to disregard irrelevant details and capture highly semantic features instead.
This approach has been successfully implemented in models like I-JEPA for \mbox{images \cite{Assran2023},} V-JEPA for \mbox{video data \cite{Bardes2024} \cite{Assran2025}}, and MC-JEPA for jointly learning motion and content features \cite{Bardes2023}.
Concepts like \ac{jepa} and \ac{ssl} naturally also find application in the automotive domain.
For instance, T-JEPA \cite{Li2024} is a model to calculate the similarity of trajectories and in \cite{Balasubramanian2022} expert-guided augmentations of traffic scenarios are used to train a self-supervised representation learning model. 

In the context of autonomous driving, safety is of the utmost importance. 
One way to improve safety is to monitor the \ac{av} by the employment of monitoring frameworks. 
These are typically online mechanisms that run in the background, observe the system behaviour and detect anomalies or safety violations during live operation. 
The scope of live operation may also be referred as \mbox{\textit{online}, \textit{run-time} \cite{Osman2019, Hashemi2021},} or \textit{operation phase} \cite{ISO21448}.
The primary objective of such frameworks is to monitor the system and provide a continuous safety assessment during the \ac{av}'s lifecycle.
In automotive applications, monitoring frameworks are also referred to as \ac{ismr} systems \cite{DeGelder2025}, which are typically embedded within high-level safety frameworks such as a \ac{saf}~\cite{DeGelder2025}.


Methods to detect anomalies or outliers often can be applied within a monitoring framework, as their underlying methodology often is similar.
There are several methods based on one sensor modality.
For instance, in \cite{Bogdoll2025}, a LiDAR-based framework to detect model failures by comparing the discrepancies of two different processing streams is proposed.  
In \cite{Shoeb2025a}, \ac{ood} LiDAR samples are detected, in order to support an adaptive neural network architecture with a continual learning setup. 
In \cite{Rivera2025}, inconsistencies between the detections of flipped LiDAR point clouds are detected for the task of active learning.
Further, the framework \acs{lsvos} \cite{Piroli2023} synthesizes outliers by adding noise in the latent space, to train an outlier detector for bounding boxes based on lidar point clouds.
There are also works that leverage the discrepancies between two object representations from two different sensor modalities to detect spoofing \mbox{attacks \cite{Alkanat2025},} or anomalies in object representations \cite{Fertig2024}.
Representation learning has proven effective for scenario-based detection. Triplet learning applied to traffic scenarios enabled detection of novel infrastructure types \cite{Wurst2021, Wurst2024}.
Further, the latent space of a variational autoencoder was used to cluster and detect anomalies in automotive data \cite{Bogdoll2023}.

Although the importance of monitoring frameworks is already well known in the field of safety, they may be limited in their application due to the need for multiple sensors, the modularity of the sensors themselves and the level at which they are applied.
The monitoring framework proposed in this work does not require multiple sensors; a single arbitrary sensor is sufficient.
Furthermore, the framework is designed to monitor the object list, which represents a critical level in \acp{av} as other planning tasks (\eg trajectory \mbox{prediction \cite{Fertig2025, Neumeier2024}}, trajectory generation \cite{Egolf2025} or scenario \mbox{clustering \cite{Rossberg2025}}) are highly dependent on the quality and accuracy of the underlying object list.

\section{Methodology} 
\label{ch:methodology}

In this section, a detailed exposition of the proposed online monitoring framework for anomaly detection is presented. 
In \Cref{ch:methodology:problem_formulation} a problem formulation is provided to illustrate the goal of the proposed framework.
As shown in \Cref{fig:fig_framwork_overview}, the framework contains three main components, namely the perception system, the embedding method, and the anomaly detection method. 
First, in \mbox{\Cref{ch:methodology:perception_system}} the perception system is described, in which \ac{mot} is applied to obtain object state estimations in the form of time series data.
Second, \mbox{in \Cref{ch:methodology:embedding}} the self-supervised embedding procedure is introduced. 
Third, in \Cref{ch:methodology:anomaly_detection} latent object embeddings are used for anomaly detection.
Furthermore, the potential domains of application and inference are addressed in \Cref{ch:methodology:inference_application}.

\subsection{Problem Formulation}
\label{ch:methodology:problem_formulation}
The main objective of this work is to develop a monitoring framework that can predict whether an object observation is anomalous. 
This task is represented by the mapping 
\begin{equation}
   E_{\theta} \circ \varphi : \bm{\mathcal{O}} \mapsto \hat{a} ,
\end{equation}
where the object encoder $ E_{\theta}$ and the anomaly detection method $\varphi$ jointly process $\bm{\mathcal{O}}$ and make a binary anomaly prediction \mbox{$\hat{a} \in \{0,1\}$}.
Hereby, $\bm{\mathcal{O}}$ is an object state representation, \ie object, represented as multi variate time series and $ E_{\theta}$ is realized using a machine learning-based sequence encoder.
The object \mbox{encoder $E_{\theta}$} is trained in a self-supervised fashion using the \ac{jepa} paradigm to translate object state representations into a latent embedding space.
The object encoder $E_{\theta}$ is the core component of the framework, as it produces informative object embeddings.
These resulting object embeddings are utilized by the anomaly detection method $\varphi$ to identify anomalous samples. 
A key advantage is the \ac{jepa}-based framework design, which does not require anomaly labels in the training or inference phase.
Therefore, the proposed framework is particularly suitable for online application, in the context of \ac{sotif} \cite{ISO21448}.

\begin{figure}    
   \vspace{-0.55cm}
   \centering
   \includegraphics[width=0.48\textwidth]{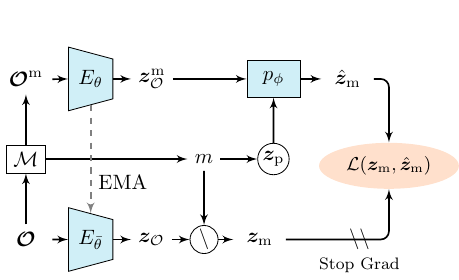}  
   \caption{
        Architecture of the self-supervised embedding method based on \ac{jepa} \cite{LeCun2022}.
        }
    \label{fig:fig_embedding_arch}
    \vspace{-0.4cm}
\end{figure}

\subsection{Perception System} 
\label{ch:methodology:perception_system}

An \ac{av} is able to perceive its immediate environment through its perception system, which usually consists of several different sensors, \eg radar, camera and \mbox{LiDAR}.
A common approach for processing raw sensor data is to apply an object detection method specific to the sensor, followed by the utilization of a tracking algorithm to capture the object's temporal progression.
This two-step process is also known as \mbox{\textit{tracking-by-detection}} scheme; the overall task it addresses is \ac{mot}.
If there are multiple sensors available, their results can be combined through a process known as sensor fusion.
The result of \ac{mot} is a list of object state representations over time, which is also referred to as object list.
Thereby, an object list  \mbox{$ \bm{L} = { \left\{\bm{ \mathcal{O}}_{i} \right\} }^{N}_{1} $} consists of $N$ objects $\bm{\mathcal{O}}_i$, with each object denoting the state estimations of a specific traffic participant.
Each object \mbox{ $ \bm{ \mathcal{O} }_i = ( \bm{\mathcal{T}}_i,\bm{\mathcal{G}}_i )  $} is a multivariate time series over the sequence length $T$ and constituted by two elements: its \mbox{trajectory $ \bm{\mathcal{T}}_i$} and its object \mbox{attributes $\bm{\mathcal{G}}_i$}.
The trajectory \mbox{$ \bm{ \mathcal{T}}_i  = ( \bm{x},\bm{y} ) \in \mathbb{R}^{2 \times T} $} contains the object's longitudinal $\bm{x}$ and lateral $\bm{y}$ position, depicted within a global coordinate system, with the EGO vehicle starting at the origin.
The object attributes \mbox{$ \bm{ \mathcal{G} } = ( \bm{v}, \bm{\psi} ) \in \mathbb{R}^{ 2 \times T}$}  contain object features as \mbox{velocity $\bm{v}$} and \mbox{orientation $\bm{\psi}$}.

For \acp{av} the quality and accuracy of the object list $\bm{L}$ from the perception system is of paramount importance, given that core driving functions are reliant on it.
To supervise $\bm{L}$ and detect anomalous objects, an online monitoring framework for object lists is proposed.

\subsection{JEPA-based Embedding Method} 
\label{ch:methodology:embedding}

An overview of the monitoring framework is provided in \Cref{fig:fig_framwork_overview} while the framework to train the object \mbox{encoder $E_{\theta}$} is illustrated in \Cref{fig:fig_embedding_arch}.
The objective of the \ac{jepa}-based embedding is to train a powerful object encoder $E_{\theta}$.
For the training procedure a self-supervised prediction task is constructed, where the latent representations of masked object timesteps are predicted.

\subsubsection{Architecture}
The architecture of the proposed \ac{jepa}-based embedding method is inspired by \cite{Assran2025} and \cite{Li2024}.
Thereby, the fundamental JEPA architecture \cite{LeCun2022} with context and target branch is adopted.
The proposed method works on object-level, therefore a sample is represented by an object $\bm{ \mathcal{O} }$.
The object encoder $E_{\theta}$ is also referred to as context encoder and shares the same architecture as the target encoder $E_{\bar{\theta}}$. 
However, the weights of both encoders are not identical, instead the weights $\bar{\theta}$ of $E_{\bar{\theta}}$ are an \ac{ema} \cite{Tarvainen2017} of the weights $\theta$ from $E_{\theta}$.
This is a standard regularization approach for \ac{jepa} in order to prevent \mbox{collapse \cite{LeCun2022, Assran2023, Bardes2023}.}
A Transformer \mbox{encoder \cite{Vaswani2017}} is used for $E_{\theta}$ and $E_{\bar{\theta}}$ to adequately  process the objects of varying sequence \mbox{length $T$}.
The predictor $p_{\phi}$ is realized via a lightweight transformer decoder \cite{Vaswani2017}.

\textbf{Masking.}
An essential component of the proposed architecture is the masking procedure $\mathcal{M}$, which is fundamental for the creation of the self-supervised prediction task.
The masking procedure $\mathcal{M}$ samples $N_{\textrm{M}}$ random indices \mbox{$m \in \mathbb{R}^{N_\text{M}}$} and masks them, by replacing the content of the corresponding timesteps with zero. 
Based on this, the masking procedure \mbox{$\mathcal{M}: \bm{\mathcal{O}} \mapsto \bm{\mathcal{O}}^{\text{m}}$} a masked object \mbox{$\bm{\mathcal{O}}^{\text{m}} \in \mathbb{R}^{T \times N_F}$} is created, which is still of same dimensionality as $\bm{\mathcal{O}}$.
Additionally, a binary masking feature \mbox{$\bm{b} \in \mathbb{R}^{T}$}, indicating if a \mbox{timestep $t$} is masked, is added to $\bm{\mathcal{O}}$ and $\bm{\mathcal{O}}^{\text{m}}$, thereby increasing their dimensionality to $\mathbb{R}^{T \times (N_F+1)}$.
The masked indices $m$ serve as prediction objective for the context branch, while the target branch produces the corresponding target embedding of $m$. 
Thereby, $\mathcal{M}$ enables the self-supervised learning procedure.

\textbf{Context Branch.}
The masked object $\bm{\mathcal{O}}^{\text{m}}$ is processed by the context encoder \mbox{$E_{\theta}: \bm{\mathcal{O}}^{\text{m}} \mapsto \bm{z}_{{\mathcal{O}}}^{\text{m}}$}, whereby \mbox{$\bm{z}_{{\mathcal{O}}}^{\text{m}} \in \mathbb{R}^{T  \times D}$} is the object embedding that consists out of $T$ embeddings tokens of \mbox{dimension $D$}, one for each timestep $t \in T$.
Subsequently, the predictor performs the mapping \mbox{$p_{\phi}: (\bm{z}_{{\mathcal{O}}}^{\text{m}},\bm{z}_{\textrm{p}}) \mapsto \hat{\bm{z}}_{\textrm{m}}$} to predict the embedding tokens \mbox{$\hat{\bm{z}}_{\textrm{m}} \in \mathbb{R}^{N_{\textrm{M}} \times D}$} of the masked timesteps of  $\bm{\mathcal{O}}^{\text{m}}$.
Hereby, \mbox{$\bm{z}_{\textrm{p}} \in \mathbb{R}^{N_{\textrm{M}} \times D}$} are the mask tokens, which consist of a learned vector fused with the positional embeddings of $m$.

\textbf{Target Branch.}
The target encoder \mbox{$E_{\bar{\theta}}: \bm{\mathcal{O}} \mapsto \bm{z}_{\mathcal{O}}$} embeds the full unmasked object $\bm{\mathcal{O}}$ and generates the target embedding $\bm{z}_{\mathcal{O}} \in \mathbb{R}^{T \times D}$.
The unmasked token indices, \ie the inverse indices of $m$, are be removed, so only $\bm{z}_{\textrm{m}} \in \mathbb{R}^{N_{\textrm{M}} \times D}$ remains, which represents the tokens that were masked in the context branch, as in \cite{Bardes2024}.

The outputs of the context and target branch, \mbox{$\hat{\bm{z}}_{\textrm{m}}$ and $\bm{z}_{\textrm{m}}$}, respectively, are used to train $E_{\theta}$ and $p_{\phi}$ and thereby shape the latent embedding space.

\subsubsection{Training Procedure and Loss}
During training, the weights $\theta$ of the context encoder $E_{\theta}$ and $\phi$ of the predictor $p_{\phi}$ are learned.
The weights $\bar{\theta}$ of the target encoder $E_{\bar{\theta}}$ are the \ac{ema} of $\theta$.
Because the predictor $p_{\phi}$ is kept small, most of the representational work is left to the larger encoder $E_{\theta}$, although the predictor still contributes to the training signal. 
During training the masking procedure $\mathcal{M}$ randomly samples and masks $N_{\textrm{M}}$ timesteps of an object $\bm{\mathcal{O}}$, in order construct the self-supervised prediction task, trained by the objective
\begin{equation} 
\arg\min\limits_{\theta, \phi,  \bm{z}_{\textrm{p}}} = \mathcal{L}(    p_{\phi}( E_{\theta}(\bm{\mathcal{O}}^{\text{m}}), \bm{z}_{\textrm{p}},) - \textrm{sg}(E_{\bar{\theta}}(\bm{\mathcal{O}})) ), 
\end{equation}
where $\textrm{sg}(\cdot)$ represents the stop-gradient operation.
As loss function $\mathcal{L}$, the L1-loss is utilized, as it enables more stable training \cite{Bardes2024}.
The goal of this training procedure is to train a powerful object encoder $E_{\theta}$, which can be used during inference to perform the mapping
\begin{equation}
    \mbox{$E_{\theta}: \bm{O} \mapsto \bm{z}_{\mathcal{O}}$}.
\end{equation}

In essence, the presented embedding method is based on \ac{jepa} \cite{LeCun2022} and adapted to create a suitable prediction task for the application of automotive time series data.
A key advantage is that the self-supervised training procedure enables the model to comprehend the automotive data and generate rich representations without the need for labels.

\subsection{Anomaly Detection} 
\label{ch:methodology:anomaly_detection}
In this section, an anomaly definition is provided and the anomaly detection procedure is introduced.

\subsubsection{Anomaly Definition}
In this work, the SOTIF standard, \mbox{ISO 21448 \cite{ISO21448}}, is used as reference, as it defines the scope of application at hand. 
In \ac{sotif} it is explicitly differentiated between \textit{known} and \textit{unknown} risks, also referred to as \ac{sotif} areas 1 \& 2 and areas 3 \& 4, respectively \cite{ISO21448}.
In the context of validating \acp{av}, it can be assumed that the data from the training and test phase can be handled by the \ac{av} and therefore is considered \textit{known}, \ie  normal (cf. \ac{sotif} areas 1 and 2) and not representing anomalous behavior of objects.
In contrast, data characteristics or behaviors that are absent or inadequately represented in the dataset are considered to be \textit{unknown} and consequently regarded as anomalies (cf. \ac{sotif} areas 3 and 4). 
When such \textit{unknown} samples that deviate from the \textit{known} and tested dataset, appear during online application in a real-world environment, the \ac{av}'s ability to handle the situation cannot be guaranteed, which underscores the importance of such monitoring mechanisms.
Following this anomaly formulation, a dedicated monitoring framework is introduced that targets the identification of abnormal object dynamics and properties, with the goal of uncovering safety-relevant irregularities that are essential for improving the operational safety of \acp{av}.

\subsubsection{Anomaly Detection Method}
The anomaly detection method $\varphi$ evaluates the object embedding $\bm{z}_{\mathcal{O}}$ to make a binary anomaly prediction \mbox{$\hat{a} \in \{0, 1\}$}.
This work aims to use established approaches for anomaly detection. 
However, as these methods require constant input values, the object embedding $\bm{z}_{\mathcal{O}}$ of the variable length $T$ is summarized.
For this, typically the max \cite{Ranasinghe_2023_ICCV} or average embedding \cite{Jose_2025_CVPR, Mukhoti_2023_CVPR} is used.
Here, the maximum embedding across the time dimension is processed by the anomaly detection method.

\begin{equation}
     \varphi : \text{max}(\bm{z}_{\mathcal{O}}) \mapsto \hat{a},
\end{equation}
to predict whetter a sample is anomalous or not.

The anomaly detection method $\varphi$ is trained for online application as follows:
The objects from the train set represent the \textit{known}, \ie normal, behavior and are processed by $E_{\theta}$ to create latent embeddings which are used to fit $\varphi$.
Through these compressed representations of normal behavior, $\varphi$ is able to capture their properties and characteristics.
For testing this framework, the test set is altered so it contains object attributes or behavior that were not present in the train set.
The embeddings of the test set are used by $\varphi$ to make an anomaly prediction $\hat{a}$ for each sample, whether it is anomalous or not.
During inference, \ie real-world application, the objects detected by perceptions system are embedded by $E_{\theta}$ into latent object embeddings, which are evaluated by $\varphi$, in order to detect \textit{unknown} object behavior.
The utilized established anomaly detection methods are described in \Cref{ch:dataset:implementation_details}.

\subsection{Inference and Application} 
\label{ch:methodology:inference_application}

The proposed monitoring framework is designed for online deployment, such as during the operation phase of \ac{sotif} \cite{ISO21448}.
Development and testing data generally represent \textit{known} system behavior, while real-world operation introduces \textit{unknown} samples that may contain novel risks or safety violations.
Detecting these anomalies is challenging due to the absence of labels; this is addressed via the self-supervised JEPA paradigm \cite{LeCun2022}, which enables label-free training.

As the framework only detects unknown states without intervening, retrospective analysis of flagged objects suffices. 
Detected anomalies can be used to build or expand datasets of challenging cases, improve the development process, or support continuous learning strategies \cite{Shoeb2025a}.
In practice, detections are often sent to the \ac{oem} for analysis \cite{Osman2019, Segler2019}, and they can also be leveraged in active learning pipelines \cite{Rivera2025} or open-world learning setups \cite{Balasubramanian2023a}. 

This work presents an online monitoring framework using \ac{jepa} embeddings to detect anomalies in automotive object lists. 
\ac{jepa}'s self-supervised training requires no anomaly labels, enabling deployment in novel environments.

\section{Dataset and Implementation Details} \label{ch:dataset}

\subsection{Dataset}
\label{ch:dataset:dataset}

To evaluate the proposed monitoring framework the real-world and publicly available nuScenes dataset \cite{Caesar2020} is used.
The nuScenes dataset is a well-established automotive dataset that provides raw data from camera, LiDAR, and radar sensors recorded from the perspective of an EGO vehicle.
The nuScenes dataset consists of $1,000$ traffic scenes, each of which is $ \SI{20}{\second}$ long and recorded at $ \SI{2}{\hertz}$.

The perception system described in \Cref{ch:methodology:perception_system} produces an object list $\bm{L}$, which is analyzed by the the subsequent components.
For this, state-of-the-art \ac{mot} methods \cite{Chen2023a} have been applied on the nuScenes raw data.
In this work, the \ac{mot} method \mbox{FocalFormer3D \cite{Chen2023a}} was employed, which is based on LiDAR data only, however the proposed framework is also suitable to evaluate object lists based on multiple sensor modalities.
The minimum object length is set to \mbox{$T_{\textrm{min}} = 8$}, resulting in \mbox{$N_{\textrm{train}} = \getNumObjectLiDARTrain$} and \mbox{$N_{\textrm{test}} = \getNumObjectLiDARTest$} objects detected by FocalFormer3D from the nuScenes train, validation, test set, respectively.

\subsection{Anomaly Setup}
\label{ch:dataset:anomaly_setup}

To the best of the author's knowledge, there is no publicly available automotive dataset containing labeled anomalies at object state level.
The following anomaly setup was therefore used to evaluate the proposed monitoring framework, in line with the anomaly definition from \Cref{ch:methodology:anomaly_detection}.

In this setup, an error model is applied to create artificial anomalies in objects, representing sporadic errors, caused by \eg the utilized sensor system or the detection and tracking methods.
This error model, is only applied on the test set $\bm{O}^{\text{test}}$ and creates anomalous objects $\bm{O}^{\text{test}}_{\text{a}}$ by corrupting a specific object feature at a given timestep.
The error on the feature $f$ is denoted as \mbox{$\vartheta_\text{f} \in \mathbb{R}$} and drawn for each object from a feature-specific normal distribution $ \mathcal{N} (\mu_{\text{f}}, \sigma_{\text{f}}^{2})$.
For instance, if a sampled error $\vartheta_{\text{v}}$ is applied to the velocity feature $\bm{v}$ at the specific \mbox{timestep $t \in \mathbb{Z^+}$,} this is formulated \mbox{as: $\bm{v}[t] + \vartheta_{\text{v}}$.}
This concept of creating anomalies is oriented on the sensor-agnostic \ac{pem} proposed in \cite{Piazzoni2024}, which also applies errors on object state level.
Furthermore, the \ac{vos} concept \cite{Piroli2023, Du2022} also uses distribution-based errors to synthesize outliers, however this is applied on latent embeddings.
Therefore, this error model generates sporadic feature errors, simulating defects on object level.
To evaluate these types of error specifically and avoid detecting cross-correlations between features, this was applied to one feature at a time.
The test set used for evaluation contains an equal split of normal objects $\bm{O}^{\text{test}}$ and their altered counterparts $\bm{O}^{\text{test}}_{\text{a}}$.
The objects from the train set $\bm{O}^{\text{train}}$ only contain normal objects and remain unchanged.

\subsection{Implementation Details}
\label{ch:dataset:implementation_details}

The codebase of the proposed method is made publicly available on GitHub\footnote{GitHub: \getGitHubURL}.
\subsubsection{Embedding Method}
The context encoder $E_{\theta}$ is realized by an encoder-only Transformer \cite{Vaswani2017} with \getEncoderNumberHead heads of depth \getEncoderDepth.
As head, a \getEncoderNumLayerMLP-layer MLP is applied.
$E_{\theta}$ \mbox{contains \getEncoderNumParams} trainable parameters.
The context encoder $E_{\bar{\theta}}$ shares the same architecture, but without any trainable parameters as an \ac{ema} of is \getEncoderEMA used.
The \mbox{predictor $p_{\phi}$ }is implemented using a transformer decoder with \getPredictorNumberHead heads of depth \getPredictorDepth.
Hereby, no head is applied and the number of trainable parameters is \getPredictorNumParams.
The latent space is of size $D = \getEncoderLatentDim$ and as optimizer ADAM is used \cite{Kingma2015}.
Because the proposed framework is intended for online use, $E_{\theta}$ and $p_{\phi}$ are designed to avoid high computational costs and require only \getNumFLOPs  FLOPs per sample.
The training procedure is executed for \getEpochsPretraining epochs using a learning rate of \getLearningRate.
During the training the masking procedure $\mathcal{M}$ \mbox{masks $N_{\textrm{M}} = \getNumMasksPretraining$} timesteps.

\subsubsection{Anomaly Detection Method}
As anomaly detection method $\varphi$ the established  \ac{abod} \cite{Kriegel2008} is used.
Furthermore, also experiments with other anomaly detection methods as \ac{lof} \cite{Breunig2000}  with \getLOFnumNeighbor neighbors or 
\ac{gmm} shows similar results, underlying the informativeness of the created latent space.
The GMM applies \getGMMComps components and makes predictions with utilizing the the log-likelihood score in tandem with an adaptive threshold based on the object embeddings of the train set.
For evaluation of the anomaly detection task the metrics \ac{auroc}, F1-score, \ac{mcc}, and accuracy are used.
Additionally, the FPR95, which reports the \ac{fpr} for an \ac{tpr} of 95\% and the TPR5 / TPR1, which measures the \ac{tpr} when the \ac{fpr} is 5\% and 1\%, find application.

\subsubsection{Baseline Approach}
To enable a comparison with the presented framework, anomaly detection identical to \Cref{ch:methodology:anomaly_detection} is performed on the objects in their original input space, \ie $\bm{\mathcal{O}}$.
The same anomaly setup as in \mbox{\Cref{ch:dataset:anomaly_setup}} is used, with the only difference being that the data was converted into a suitable format.
Thereby, the difference between the original object space $\bm{\mathcal{O}}$ and the created latent representation space $\bm{z}_{\mathcal{O}}$ should be illustrated.

\section{Experiments and Results}
\label{ch:experiments_results}

This section evaluates the proposed monitoring framework by addressing the following questions:
\begin{enumerate}
    \item \textit{Is it possible to detect anomalies using the latent embeddings of object state representations?}
    \item \textit{How reliable is the anomaly detection under the strict false-alarm budget of \acp{av}?}
    \item \textit{What effect does the severity of the anomaly have?}
\end{enumerate}

\begin{table*}[!tp] 
    \vspace{0.2cm}
    \centering
    \caption{Anomaly Detection Performance. The arrows indicate the optimal is high or low, the best results are indicated in bold, and mean $\pm$ standard deviation across $n=\getNumbRunsAnomalyDetectionPerformance$ training runs of $E_{\theta}$ are shown.} 
    \label{tab:ad_performance}
    \includegraphics[width=0.95\textwidth]{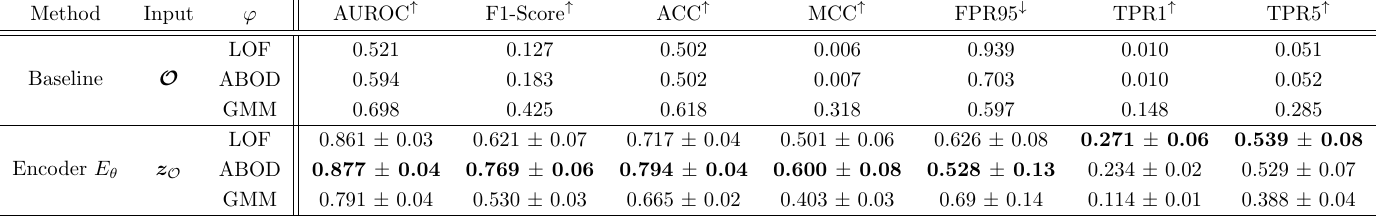}  
    \vspace{-0.3cm}
\end{table*}

\vspace{-0.3cm}
\subsection{Anomaly Detection Performance}
\label{ch:experiments_ad_performance}
To address the first research question, in \Cref{tab:ad_performance} the anomaly detection performance of the proposed monitoring framework is illustrated and compared to a baseline approach.
The monitoring framework was trained on the normal objects $\bm{O}^{\text{train}}$ from the nuScenes train split.
For evaluation, anomalous objects $\bm{\mathcal{O}}^{\text{test}}_{\text{a}} $ are created using the error model described in \Cref{ch:dataset:anomaly_setup}, altering the velocity feature based on $\mu_{\text{v}} = \SI{5}{m/s}$ and $\sigma_{\text{v}}=\SI{0.1}{m/s}$.
The test set contains an equal number of normal objects $\bm{\mathcal{O}}^{\text{test}}$ and anomalous objects $\bm{\mathcal{O}}^{\text{test}}_{\text{a}}$ that have been altered by the error model.

The results presented in \Cref{tab:ad_performance} demonstrate the effectiveness of the proposed monitoring framework in detecting sporadic anomalies in objects.
It can be seen, that the baseline approach is not able to solve the task using the original feature space  $\bm{\mathcal{O}}$, while the anomaly detection methods perform considerably better when utilizing the object embeddings $\bm{z}_{\mathcal{O}}$.
This comparison demonstrates the effect of $E_{\theta}$ and the suitability of the created latent embedding space for the task of anomaly detection,  thereby providing a positive response to the first research question.

\subsection{Reliability of Anomaly Detection}
\label{ch:experiments_ad_reliability}
\begin{figure}
     \centering
     \begin{subfigure}[b]{0.234\textwidth}
         \centering
         \includegraphics[width=\textwidth]{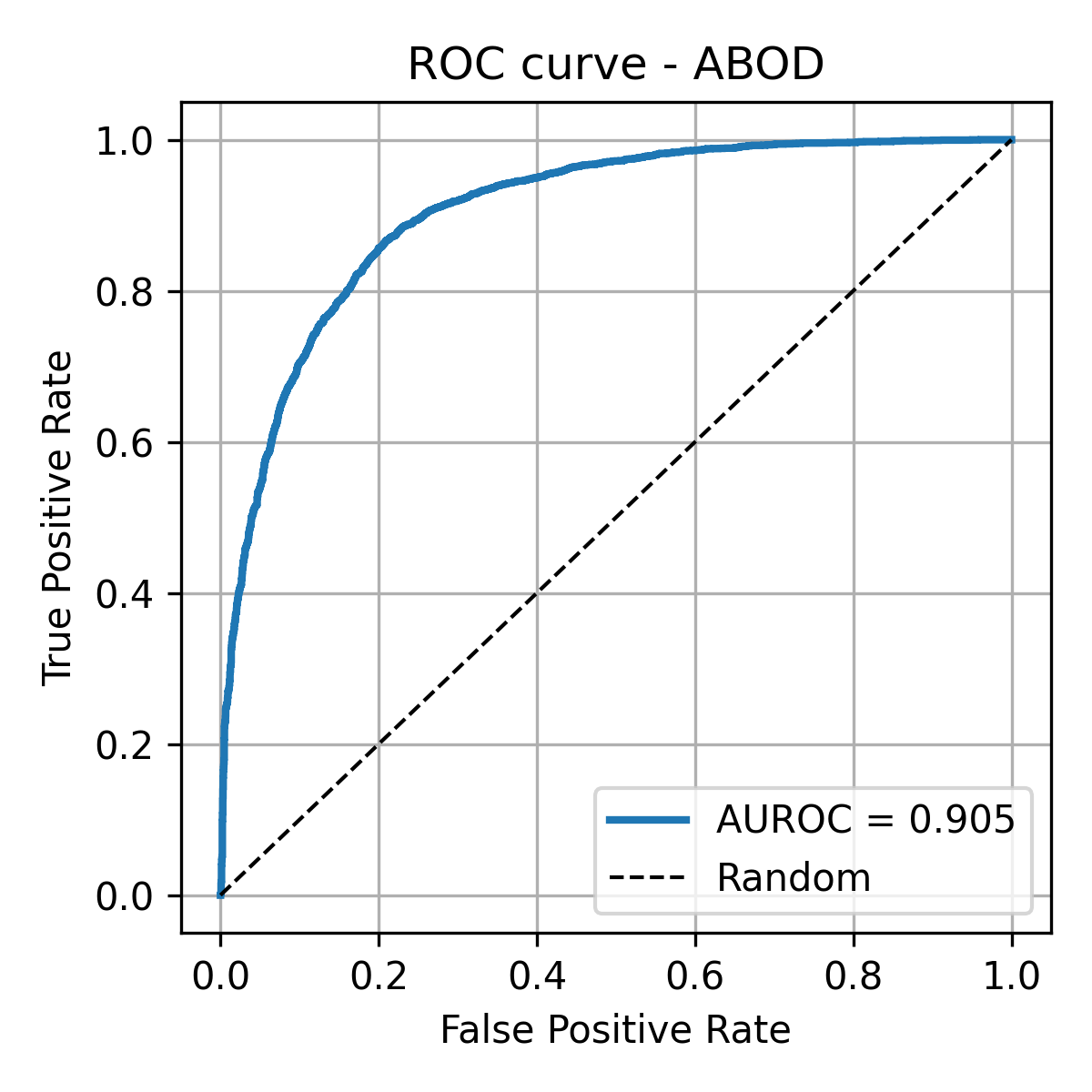}
         \caption{ROC-Curve}
         \label{fig:fig3_rov_curve}
     \end{subfigure}
     \begin{subfigure}[b]{0.234\textwidth}
         \centering
         \includegraphics[width=\textwidth]{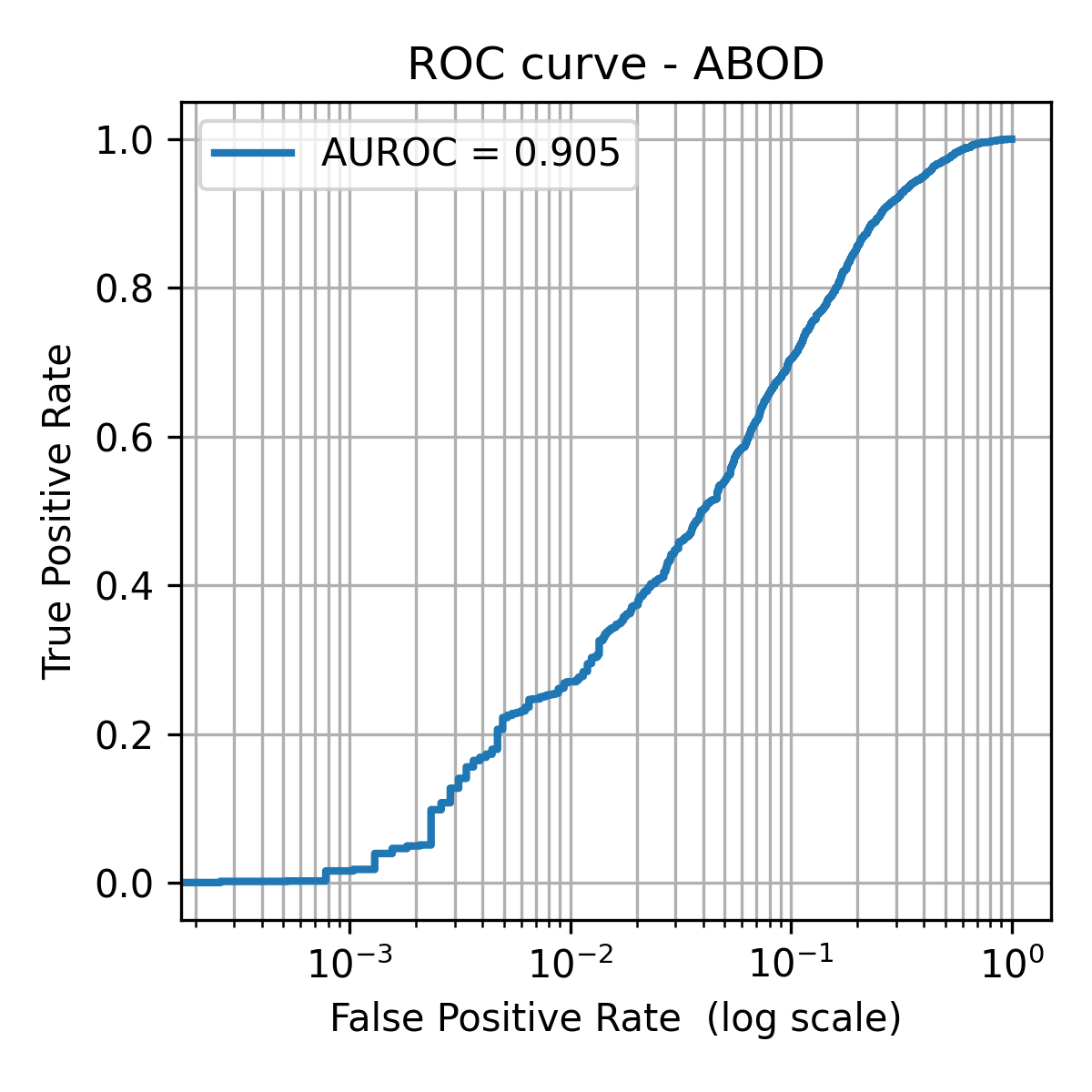}
         \caption{ROC-Curve with log-scale}
         \label{fig:fig3_rov_curve_logscale}
     \end{subfigure}
     
     \caption{
            Plot of two variants of the \ac{roc} curve, from a single run using \ac{abod} as $\varphi$.   
        }

    \label{fig:fig3_roc_curves}
    \vspace{-0.5cm}
\end{figure}

A crucial characteristic of any monitoring framework is its reliability, as processing and analyzing false alarms, \ie false positives, is expensive and therefore only possible within certain budgetary limits, also known as false alarm budgets.
To answer the second research question and quantify the framework's reliability, the same setup as in the first research question is used.
Thereby, the \ac{roc} curve for \ac{abod} is shown in \Cref{fig:fig3_roc_curves} and three characteristic operating points of the \ac{roc} curve are highlighted.
An established metric is the  FPR95, which represents the \ac{fpr} when the \ac{tpr} is 95\%, for \ac{abod} the FPR95 is at 52.8\%.
Furthermore, TPR1 and TPR5 provide the \ac{tpr} for a \ac{fpr} of 1\% and 5\%, respectively.
For a monitoring framework to be applied online within a fleet of \acp{av}, a low \ac{fpr} is essential, otherwise not all detections could be processed due to a limited false alarm budget.
Using \ac{abod} as anomaly detection method, even for a \ac{fpr} of 1\% still 23.4\% anomalies are detected, for a \ac{fpr} of 5\% it is 52.9\%.
\Cref{fig:fig3_rov_curve_logscale} specifically highlights this area of the \ac{roc} curve.
Therefore, it is concluded that the framework remains functional while staying within a strict false alarm budget.

These quantitative values are also presented in \Cref{tab:ad_performance}.
Overall, the proposed framework is capable of delivering reliable results and perform well even with a low FPR.

\subsection{Effect of the Anomaly Severity} 
\label{ch:experiments_anomaly_severity}

\begin{figure}[t!]
   \centering
   \includegraphics[width=0.4\textwidth]{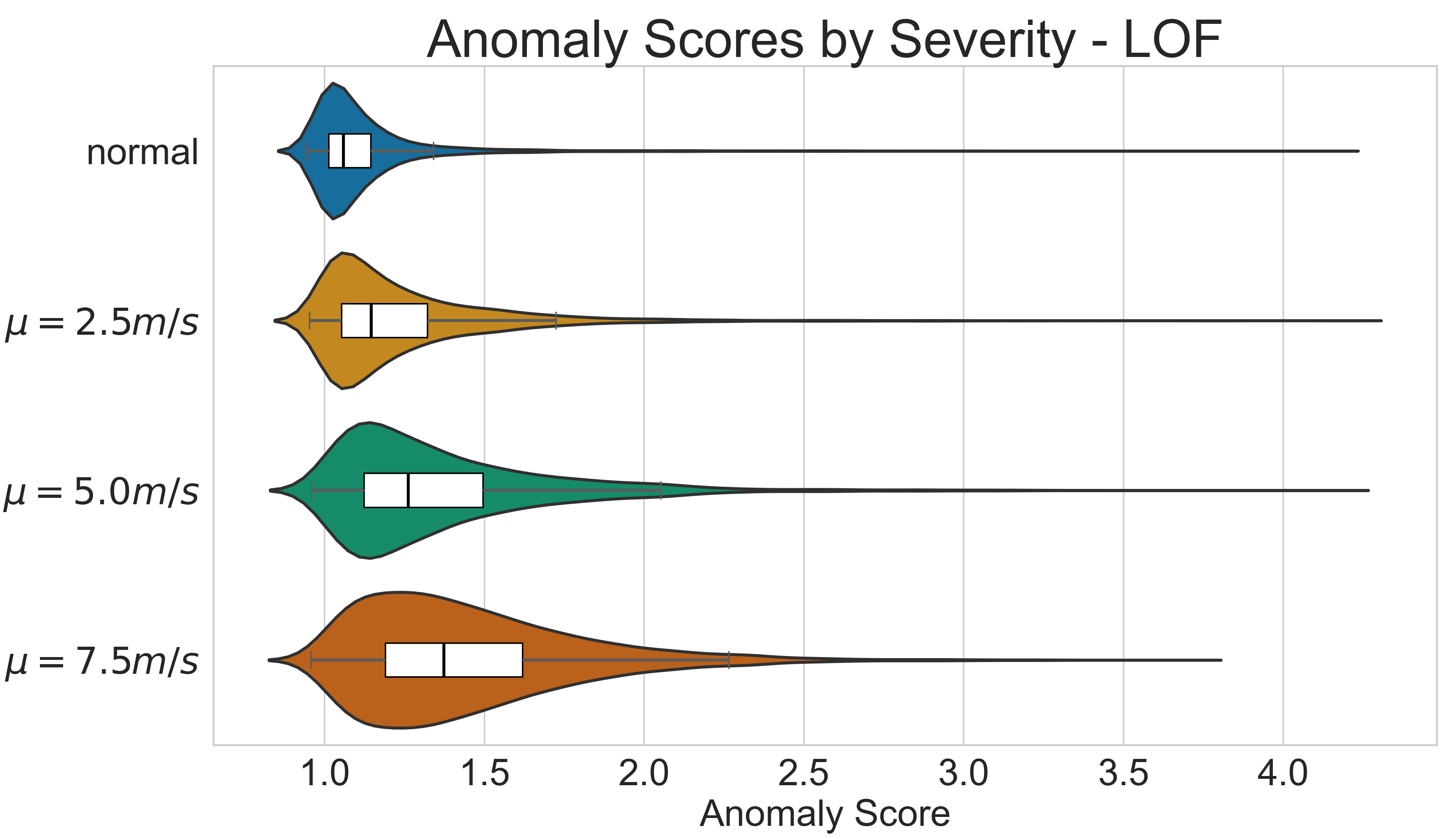}
   \caption{Visualization of the anomaly scores from the \ac{lof} for normal objects and altered objects using different means $\mu_{\text{v}} \in \{2.5, 5, 7.5\} $ $\SI{ }{m/s}$.}
    \label{fig:fig_anomaly_severity}
    \vspace{-0.4cm}
\end{figure}

To address the third research question, the setup from \Cref{ch:experiments_ad_performance} was utilized with different values for $\mu_{\text{v}}$, which represent the severity of the created anomaly.
\Cref{fig:fig_anomaly_severity} illustrates the anomaly scores from the \ac{lof} for the normal objects and anomalous objects of three severities.
The higher the score, the more anomalous the sample, with normal scores tending to be close to 1.
As $\mu_{\text{v}}$ increases, it can be seen that the distribution of anomaly scores shifts to the right.
This is logical, as higher anomaly severity should result in a higher anomaly score.

\section{Conclusion}
\label{ch_conclusion}

This paper presents an online monitoring framework with the objective of continuously monitoring the \ac{av}'s object list and detecting anomalies at the level of object state representations. 
The object list is a key interface for autonomous vehicles, and its integrity is essential for the overall performance and safety of \acp{av}.
The framework consists of three main components.
First, the perception system that observes the environment and produces the object list using an arbitrary or multiple sensors.
Second, the self-supervised embedding method, which encodes the objects into an expressive latent representation space.
Hereby, a \mbox{\ac{jepa}-based \cite{LeCun2022}} training procedure is applied to train a strong object encoder.
Third, the created \ac{jepa} embeddings are evaluated by an anomaly detection method, in order to detect anomalies on object state level.
A key advantage of the proposed framework is that neither the embedding method nor the anomaly detection method require anomaly labels for training.
This is particularly useful when the system is deployed in a real-world setting, where new or unknown anomalies can be encountered for which in general no labels are available.
The proposed monitoring framework can be applied in an online fashion, \eg the operation phase of \ac{sotif} \cite{ISO21448}, thereby supporting the safety and validation process of \acp{av}.
The suitability of the proposed monitoring framework is demonstrated by experiments on the publicly available real-world nuScenes dataset.

\section*{Acknowledgment} 
\label{Sec:acknowledgement}
The authors acknowledge the support of ZF Friedrichshafen AG.


\bibliographystyle{IEEEtran}
\bibliography{20260130_ZoteroLib}

\end{document}